\documentclass[10pt,twocolumn,letterpaper]{article}

\usepackage{cvpr}
\usepackage{times}
\usepackage{epsfig}
\usepackage{graphicx}
\usepackage{amsmath}
\usepackage{amssymb}
\usepackage{multirow}
\usepackage{commath}
\usepackage{bm}
\usepackage{subcaption}
\usepackage{array}
\usepackage{graphicx}
\usepackage{tabularx}
\usepackage{enumitem}

\DeclareMathOperator*{\argmax}{argmax}

\newcommand\inner[2]{\langle #1, #2 \rangle}

\newcolumntype{s}{>{\centering\arraybackslash}X}

\usepackage[pagebackref=true,breaklinks=true,letterpaper=true,colorlinks,bookmarks=false]{hyperref}

\cvprfinalcopy 

\def\cvprPaperID{2309} 

\begin{document}

\title{Learning to discover and localize visual objects with open vocabulary}

\author{Keren Ye$^{1}$~~ Mingda Zhang$^{1}$~~ Wei Li$^{2}$~~ Danfeng Qin$^{2}$~~ Adriana Kovashka$^{1}$~~ Jesse Berent$^{2}$\\
$^{1}$Department of Computer Science, University of Pittsburgh, Pittsburgh PA, USA\\
$^{2}$Google Research, Zurich, Switzerland\\
{\tt\small \{yekeren, mzhang, kovashka\}@cs.pitt.edu ~~ lwthucs@gmail.com ~~ \{qind, jberent\}@google.com}
}

\maketitle

\begin{abstract}
To alleviate the cost of obtaining 
accurate bounding boxes for training today's state-of-the-art object detection models, recent weakly supervised detection work has proposed techniques to learn from image-level labels.
However, requiring discrete image-level labels is both restrictive and suboptimal.
Real-world ``supervision'' usually
consists of more unstructured text, such as captions. In this work we
learn association maps between images and captions. We then use a novel objectness criterion to rank the resulting candidate boxes, such that high-ranking boxes have strong gradients along all edges. Thus, we can detect objects beyond a fixed object category vocabulary, if those objects are frequent and distinctive enough. We show that our objectness criterion improves the proposed bounding boxes in relation to prior weakly supervised detection methods.
Further, we show encouraging results on object detection from image-level captions only. 
\end{abstract}

\section{Introduction}
\label{sec:intro}

Learning to localize and classify visual objects is a fundamental problem in computer vision.
It has a wide range of applications, including robotics, autonomous driving, intelligent video surveillance, and augmented reality.

Since the renaissance of deep neural networks,
the field of object detection has been revolutionized by a series of groundbreaking works, including RCNN \cite{girshick2016region}, Fast-RCNN \cite{Girshick_2015_ICCV}, Faster-RCNN \cite{ren2015faster} and Mask-RCNN \cite{He_2017_ICCV}.
State-of-the-art detection performance on Pascal VOC 2012 has improved from a mean average precision of less than $20\%$ in 2007 to over $80\%$ in 2018 \cite{liu2018deep}.
Today, object detectors can be run in real time on mobile devices, and self-driving cars are close to fully autonomous operation in a few cities.
However, object detection is still far from being a solved problem.
Detectors can only recognize objects from a limited vocabulary but are blind to everything else.
Due to the expense of human annotation, the largest datasets~\cite{openimages} currently available only contain hundreds of categories of objects, while humans can recognize hundreds of thousands.
It is critical to scale up the variety of our datasets, but providing extensive box or object annotations for all domains of interest is infeasible. 

\begin{figure}
    \centering
    \includegraphics[width=1\linewidth]{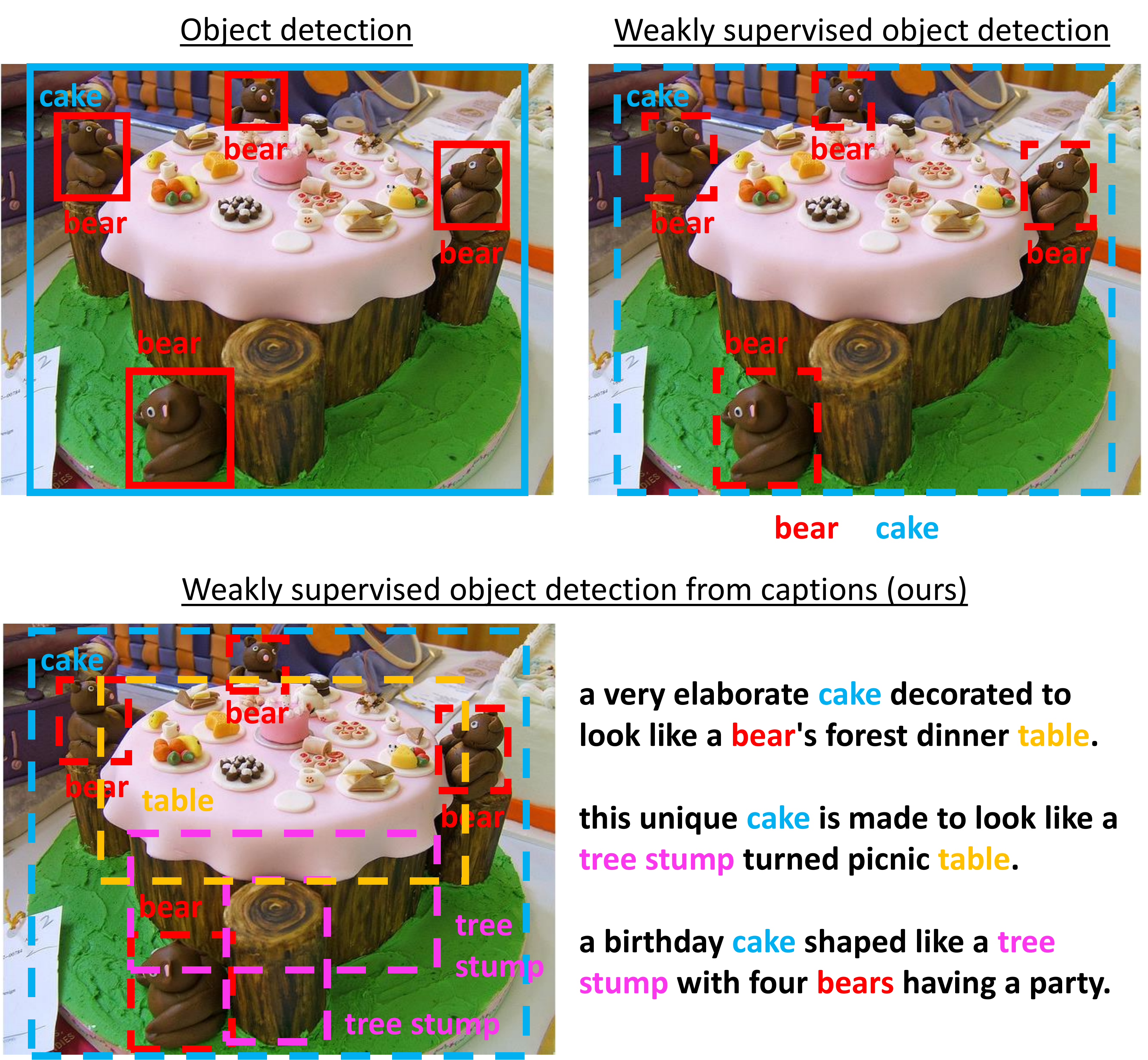}
    \caption{Existing approaches to object detection assume discrete object labels on either the box level, or the image level (in the weakly supervised setting). In contrast, our approach only requires unsegmented captions. This allows it to learn models for objects that lie beyond the fixed object category vocabulary that is provided in recent datasets, and to mine for and detect additional objects (such as ``tree stump'' and ``table'').}
    \label{fig:concept}
\end{figure}

We believe an essential step to scale up to millions of object classes is to use abundant and labor-free web data.
One pioneering work is from Chen et al. \cite{Chen_2017_CVPR}  which learns to discover and localize new objects from documentary videos by associating subtitles to video tracklets.
There is also work to associate phrases in the caption to its visually depicted objects in the image \cite{rohrbach2016grounding,hu2016segmentation}.
These show encouraging results of using noisy and weakly annotated multi-modal data for learning visual concepts and localizing objects.
However, \cite{Chen_2017_CVPR} requires video tracklets as input which is limiting, and \cite{rohrbach2016grounding} requires image captions at prediction time as well, and thereby can not be used as an object detector.

In this work, we go one step further.
We propose to learn an object detector from unconstrained image-caption pairs, and apply it to arbitrary images at inference time without any additional requirement.
The available amount of data for our problem setup is enormous.
There are tens of millions of photos uploaded to Instagram everyday, and a majority of them have titles, tags, or descriptions.
Abundant videos with subtitles are similarly available on Youtube.

Learning to localize objects using noisy image captions is challenging. 
As illustrated in Fig.~\ref{fig:concept}, free form text in captions might contain words that don't correspond to any visual objects like ``elaborate'', words referring to more than one instance in the image like ``bear'', or concepts for which no box can be provided, like ``party'' and ``forest''.

To resolve the spatial, visual, and semantic ambiguity, we propose to learn a spatial activation map to model the correlation between pixels and words. 
This map is formulated as pair-wise similarity between regions in the image and words in the caption, in the learned embedding space.
By aggregating activation weights across all regions and words, we can use a scalar to measure the likelihood that  the image and caption are paired.
Then we learn all parameters through a triplet loss which forces the similarity of an image paired with its own caption to be larger than a randomly sampled caption by some margin.
Secondly, we learn to localize objects for each class by finding bounding boxes that have a strong gradient in the activation map, across all edges of the box.
Finally, we use the generated boxes and class labels to train a Faster-RCNN style object detector but replace the softmax classification loss with a more noise-robust Multiple Instance Learning (MIL) loss.

Fig.~\ref{fig:concept} contrasts our approach from prior weakly supervised detection work. At the top we show standard object detection that requires bounding-boxes at training time (shown with a solid boundary). We also show weakly supervised object detection (WSOD) which only requires image-level object annotations (e.g. ``bear'', ``cake'') but can still \emph{predict} bounding boxes at test time (shown with a dashed boundary). In contrast, our proposed method (bottom of figure) requires no bounding boxes and no restriction of annotations to a fixed vocabulary. Instead, it can utilize natural-language captions, and learn to map frequent words in those captions to regions in the image. As a result, it can learn to detect a larger number of objects; in this case, it can localize ``tree stump'' and ``table'' in addition to ``bear'' and ``cake''.  

We evaluate both the box proposal and detection steps of our work. In terms of proposals, our objectness metric outperforms two weakly supervised methods. 
We achieve promising results in the very challenging scenario of detecting COCO objects from noisy captions.

\section{Related Work}
\label{sec:related}

There is growing interest in using large-scale data with weak supervision for computer vision tasks. In the object detection and localization tasks, gathering box annotations is even more time and labor consuming than image-level labeling. Therefore, learning detectors from weak supervision has received sustained interest \cite{deselaers2010localizing,pandey2011scene,russakovsky2012object,siva2011weakly,gokberk2014multi,wang2014weakly}.

\vspace{-0.5cm}
\paragraph{Learning from image labels}
Large-scale image classification datasets have been built \cite{imagenet_cvpr09, everingham2015pascal} where images are labeled with image classes from pre-defined dictionaries. Those labels describe what objects are in the image, but not where; yet could be used to learn to localize objects. \cite{Oquab_2015_CVPR, Zhou_2016_CVPR} proposed Global Average (Max) Pooling layer to learn class activation maps. 
The work that is most similar to ours is \cite{Diba_2017_CVPR, Wei_2018_ECCV}, which extend the Global Average Pooling work. They develop a unified three-stages weakly supervised object detection pipeline, involving (1) learning of the class activation map, (2) segmentation for refining activation map boundaries, and (3) learning of the weakly supervised object detection model. Our model differs from theirs in that we learn the class activation map from nosier data of paired images and captions, without a predefined object vocabulary. 
We also use a stricter criterion for objectness, which results in more precise boxes.

\vspace{-0.5cm}
\paragraph{Learning from text}
Lots of images come with text descriptions from the web. Attention models are commonly used in learning from text-image domains like captioning and visual question answering. For example, \cite{Anderson_2018_CVPR, Teney_2018_CVPR, Krause_2017_CVPR} use bottom-up attention to assign weights to different image patches, and
\cite{Ye_2018_ECCV} uses both bottom-up image attention and word attention to compute the similarity between two modalities. 
\cite{rohrbach2016grounding} ground phrases in images, using a reconstruction loss: the method should be able to reconstruct the phrase using a careful selection of boxes to attend to. However, this work does not enable an object detection model since it requires phrases at test time.

\vspace{-0.5cm}
\paragraph{Learning from other modalities}
There are also multi-modal learning models which localize visual objects from other modalities like audio and video. For example, \cite{Harwath_2018_ECCV} jointly discover visual objects and spoken words, and show promising localization in the image for particular spoken content. \cite{Arandjelovic_2018_ECCV, Gao_2018_ECCV} focus on separating distinguishable audio and video objects simultaneously.
\cite{Chen_2017_CVPR} learn to associate tracklets with words in documentary subtitles.
Most of these multi-modal methods primarily focus on captioning or retrieval tasks, while our main focus is localization. 
While some of these methods do report localization performance, many assume a different scenario than ours:  \cite{chen2015microsoft} assume the presence of tracklets, i.e. many views of the same object instance, which makes the problem simpler.

\vspace{-0.5cm}
\paragraph{Unsupervised proposal methods}
There are several prior methods that exploit the boundary, appearance and shape to detect generic objects. For example, Edge Boxes \cite{zitnick2014edge} measures the number of edges inside of a potential box and those that may be contours overlapping the box's boundary.
The Multiscale Combinatorial Grouping \cite{Arbelaez_2014_CVPR} builds a multiscale hierarchical segmentation contour map. Both of the above two methods depend on the traditional structured forest contours \cite{Dollar_2013_ICCV} approach. More recently, the Convolutional Oriented Boundaries \cite{maninis2016convolutional} uses a CNN to generate the multiscale oriented contours. However, all of these contour based methods only exploit the correlation between low-level edge information and objectness, and rarely utilize semantic information.

Compared to all the above methods, we are proposing a general detection framework from unconstrained free-form captions. Our model copes with the challenges of learning from noisy and weak supervision, to benefit from real-world, large-scale data.

\section{Approach}
\label{sec:approach}

\begin{figure*}[h]
    \centering
    \includegraphics[width=\linewidth]{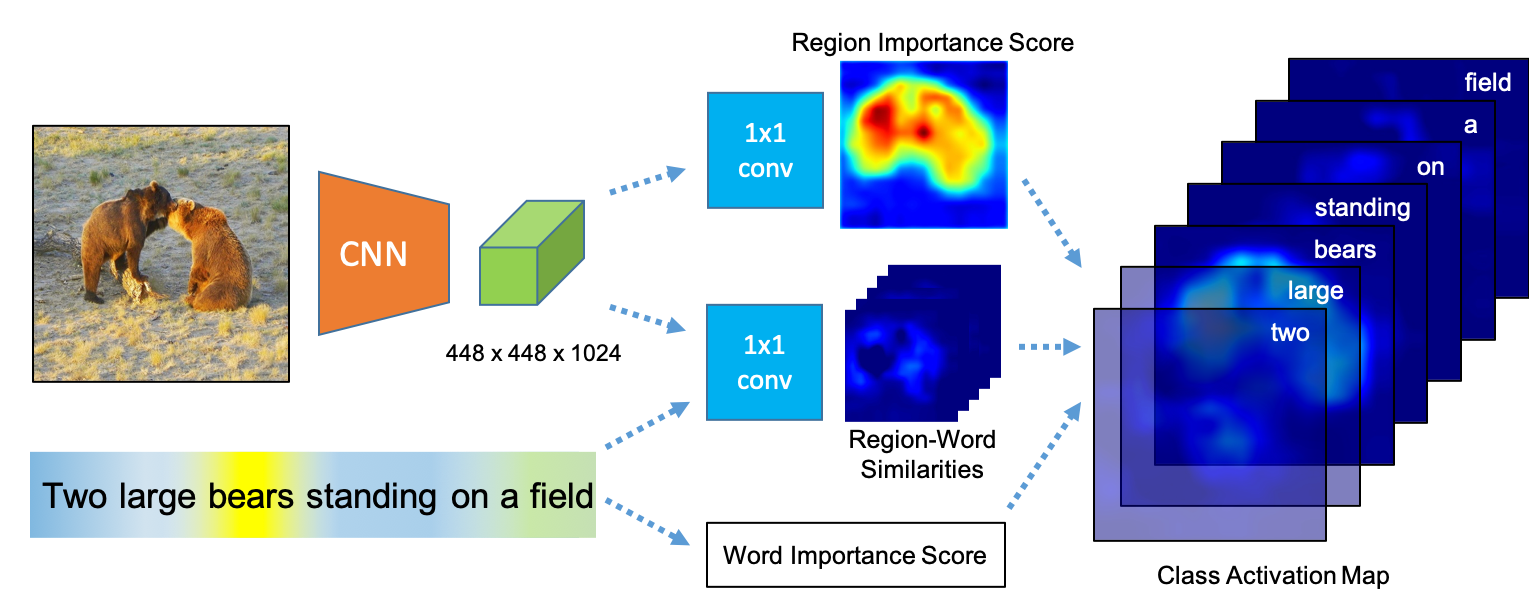}
    \caption{\textbf{Our proposed model for learning class-aware activation map.} We first compute the pair-wise region-word similarities (middle branch). Then we weigh within the image plane using the region importance score (top branch) and weigh among the words using the word importance score (bottom branch). The resulting class activation map benefits objects mining in that pixel level importance can be retrieved by the model given any \textbf{arbitrary} words.}
    \label{fig:approach}
\end{figure*}

\paragraph{Problem definition.} 

At training time, we are given pairs of images with captions, $\{\bm{x}_i, \bm{t}_i\}, i \in \{1, \dots, N\}$ where $\bm{x}_i$ denotes an image and $\bm{t}_i$ denotes a corresponding caption. Our method receives a set of region proposals $\{b(\bm{x}_i)_j\}, j \in \{1, \dots, B\}$ and ranks them using our proposed measure of objectness. 
We would like to learn an object detection model 
which takes an arbitrary image as input, and outputs a list of boxes and class labels $\{(b_1, c_1), ..., (b_K, c_K)\}$.


Note that the setup of our problem is fundamentally different from weakly supervised object detection (WSOD).
First, we do not assume that the words appearing in the caption must be associated with at least one visual object in the image. 
Second, we do not use a small pre-defined vocabulary about visual objects. 
Our problem is different from phrase-to-region studies as well,
since our model does not require a caption at prediction time.
These much relaxed requirements on the training data allow our method to work directly with a massive amount of data already on the internet:
user-uploaded photos on Instagram or Flickr along with their captions or tags,
Youtube video frames along with their subtitles,
images on webpages with their associated text, and many more.

We learn our weakly supervised object detection model in three stages. In the first stage, a multi-modal image-caption association learning method (Sec.~\ref{sec:multimodal-model}) is applied to get class activation maps (CAMs). 
In the second stage (Sec.~\ref{sec:pseudo_gt}), we transform the CAMs into a set of boxes ranked by quality, using a strict criterion over gradients inside/outside these boxes.
In the third stage, top-ranked boxes are used as pseudo-ground-truth locations, and multiple instance learning (Sec.~\ref{sec:mil}) is employed to guide a Faster-RCNN-like training pipeline.
The final object detection result is a set of boxes hypothesized to contain each of a set of object names.

\subsection{Learning the class activation map}
\label{sec:multimodal-model}

We wish to compute a map over the image for each word in the vocabulary of captions. The map measures the association between each pixel in the image and the corresponding word. 
Given input images $\bm{x}_i \in \mathbb{R}^{p \times q \times a}$ (where $p$, $q$, and $a$ are the height, width and number of channels of the image, respectively), and paired captions $\bm{t}_i$, our model learns class-aware activation maps $h(\bm{x}, c)\in \mathbb{R}^{p \times q}$, where $c$ is a class/query, $c \in V$, a vocabulary of words in captions we encounter at training time. 
We expect to see high values in parts of the map that strongly correlate with the query word. The overall approach is briefly illustrated in Figure~\ref{fig:approach}.


Given the input image $\bm{x}$ and caption $\bm{t}$, our model estimates the pairwise similarity of image regions and words, weighted by scores measuring the importance of each region and each word:

\begin{itemize}

    \item Similarity of individual region-word pairs $Sim^{ind}(\bm{x}, \bm{t})\in \mathbb{R}^{n\times n\times l}$: The pairwise similarity is calculated between each region (among $n \times n$ total regions in the image) and each word (among $l$ words in the caption). 
    We compute a representation $G^{txt}(t_j)$ for each word, and project the region representations $\bm{f}_i$ into the same space as the word representations for $t_j$, using triplet loss as described below.   
    Then we use Eq.~\ref{eq:cosine_similarity}, where the $\inner{\cdot}{\cdot}$ denotes the \emph{inner product} operation and $\| \cdot \|_2$ denotes L2 norm.
    \begin{equation} \label{eq:cosine_similarity}
        Sim^{ind}(\bm{x}_i, t_j)=\frac{\inner{G^{img}(\bm{f}_i)} {G^{txt}(t_j)}}{\|G^{img}(\bm{f}_i)\|_2\|G^{txt}(t_j)\|_2}
    \end{equation}
    where $i=1, \dots, n \times n$ and $j = 1, \dots, l$. 
    
    \item Aggregate image-caption similarity $Sim^{agr}(\bm{x}, \bm{t}) \in \mathbb{R}$:  measures how close are an entire image $\bm{x}$ and a caption $\bm{t}$ composed of words, in the learned feature space. The $Sim^{agr}$ function weighs the similarity scores $Sim^{ind}(\bm{x}, \bm{t})$ using the importance scores $S^{img}(\bm{x})$ and $S^{txt}(\bm{t})$ described below. Eq.~\ref{eq:image_caption_similarity} denotes the procedure of weighted aggregation. Note that $S^{img}(\bm{x}) S^{txt}(\bm{t})^{T}$ is an outer product (a matrix), whose values sum to 1, and $-1<Sim^{ind}(x_i, t_j)<1$. The $\odot$ denotes \emph{point-wise multiplication}.
    \begin{equation} \label{eq:image_caption_similarity}
        Sim^{agr}(\bm{x}, \bm{t})=\sum [S^{img}(\bm{x}) S^{txt}(\bm{t})^{T} 
        \odot Sim^{ind}(\bm{x}, \bm{t})]
    \end{equation}
    
    \item Region importance score $S^{img}(\bm{x})\in \mathbb{R}^{n\times n}$: We take a pre-trained CNN to extract a feature map before the last pooling layer as representation, resulting in $\bm{f} = CNN(\bm{x}) \in \mathbb{R}^{n \times n \times d}$ where $n$ is the number of evenly distributed grids and $d$ is the number of channels in this layer. Then a $1\times 1$ convolution is applied such that each region is linearly projected into one score. A softmax is used to normalize the scores across $n \times n$ regions.
    Parameters include $\bm{w}^{img}$ and $\bm{b}^{img}$; we described how we obtain them below. 
    \begin{equation} \label{eq:image_saliency}
        S^{img}(\bm{x})=\text{softmax}(\bm{w}^{img} \cdot \bm{f} + \bm{b}^{img})
    \end{equation}
    
    \item Word importance score $S^{txt}(\bm{t}) \in \mathbb{R}^l$: The caption $\bm{t}$ is treated as a sequence of words $t_1,\dots,t_l$,  where $l$ is the length. Each word $t_i$ is embedded by function into a fixed length vector $G^{txt}(t_i)$. The embeddings are then linearly projected into singular scores (one per word), which are later normalized by a softmax function within the caption. Parameters include $\bm{w}^{txt}$, $\bm{b}^{txt}$, and the weights in word embedding function $G^{txt}$.
    \begin{equation} \label{eq:word_saliency}
        S^{txt}(\bm{t})=\text{softmax}(\bm{w}^{txt}G^{txt}(\bm{t}) + \bm{b}^{txt})
    \end{equation}
    
\end{itemize}

\vspace{-0.5cm}
\paragraph{Learning.} 
To learn all terms above, we use the triplet learning framework. In particular, we require that the similarity of the paired images-captions (i.e. ones appearing paired in the data) has to be higher than that of the unpaired ones. Eq.~\ref{eq:triplet_loss} formulates the corresponding loss where $\bm{t}'$ is the negative caption sampled using semi-hard mining \cite{Schroff_2015_CVPR} and $\alpha$ defines the triplet margin. $\theta$ involves all of the parameters in the model, including $\bm{w}^{img}$, $\bm{b}^{img}$, $\bm{w}^{txt}$, $\bm{b}^{txt}$, and the embedding weights in the functions $G^{img}$ and $G^{txt}$. 
\begin{equation} \label{eq:triplet_loss}
    L(\theta)=\sum \left[Sim^{agr}(\bm{x}, \bm{t}') - Sim^{agr}(\bm{x}, \bm{t}) + \alpha\right]_{+}
\end{equation}

\vspace{-0.5cm}
\paragraph{Final output.} 
Finally, the generated class-aware activation map is computed using Eq.~\ref{eq:activation_map} where the function ``$\text{Resize}$'' resizes the $n\times n$ feature map to the same size as the original image $\bm{x}$ (using bilinear interporlation and Gaussian smoothing with kernel size of 32).
It is worth noting that for a given word $c$, the first factor $Sim^{ind}$ gives an $n \times n$ similarity map, representing the similarity between each region and the given category $c$. In the meantime, the second factor (after $\odot$) denotes the region importance regardless of the category $c$.
\begin{equation} \label{eq:activation_map}
    h(\bm{x}, c)=\text{Resize}(Sim^{ind}(\bm{x}, c) \odot (\bm{w}^{img} * \bm{f} +\bm{b}^{img}))
\end{equation}

\vspace{-0.5cm}
\paragraph{Selecting an object vocabulary.} 
Above we explain how to compute a map for the similarity of an image and a word. In practice, we only compute this similarity for a set of words (which need not be restricted to the set of object categories in an existing dataset).
To construct a meaningful object vocabulary $C$ which contains the $k^{cls}$ most distinctive words in the captioning vocabulary $V$, we use Eq.~\ref{eq:mining_objects}. 
\begin{equation} \label{eq:mining_objects}
    C= \argmax_{c\in C', C' \subset V, |C'|=k^{cls}} \bm{w}^{txt}G^{txt}(c)+\bm{b}^{txt}
\end{equation}

\vspace{-0.5cm}
\paragraph{Differences with prior work.} 
To compute $h(\mathbf{x}, c)$, we use an extension of global pooling for our multi-modal learning setting.  
A global pooling layer was used in \cite{Oquab_2015_CVPR, Zhou_2016_CVPR,Diba_2017_CVPR, Wei_2018_ECCV} in conjunction with a classification loss. 
However, in our case, supervision is in the form of captions which contain unstructured and some potentially irrelevant information.
Thus, we use triplet ranking loss and aggregate similarity scores of region-word pairs, to obtain the image-caption similarity. 
As a result, the class label can be chosen from a much larger captioning vocabulary. 
Some prior work \cite{Ye_2018_ECCV, Harwath_2018_ECCV, Arandjelovic_2018_ECCV, Gao_2018_ECCV} uses a triplet loss for multi-modal learning as well, but the goal in is retrieval, not localization.





\subsection{Generating bounding boxes}
\label{sec:pseudo_gt}

Thus far, we described how to compute a class activation map, showing which parts of an image most strongly correlate with a caption/word. We now use this map to compute objectness scores for a set of candidate boxes in the image.

\begin{figure}[t]
    \centering
    \includegraphics[width=1.0\linewidth]{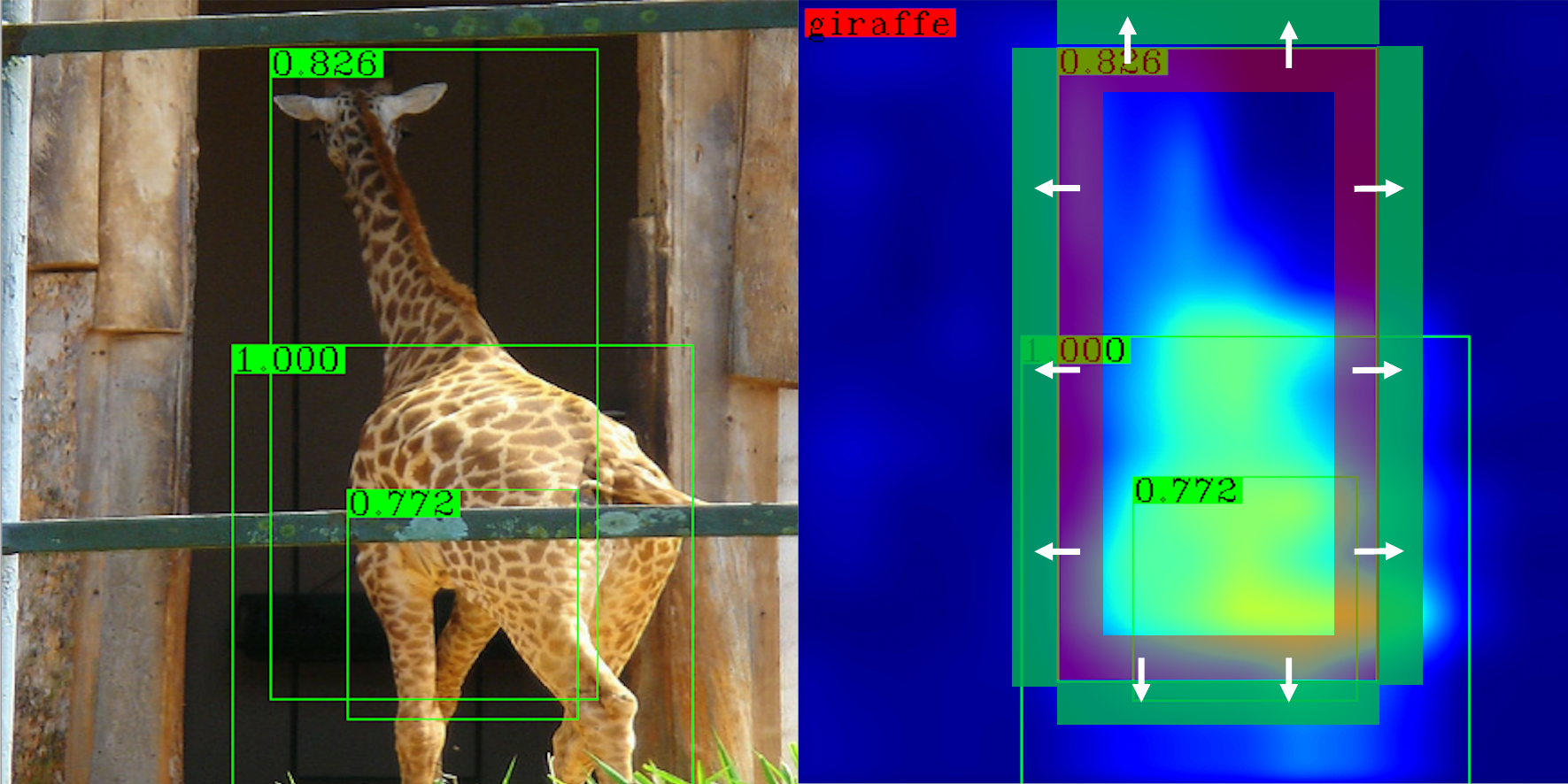}
    \caption{\textbf{Objectness score computation}. We show how to compute the score for the tightest box (which scored $0.826$). Based on the integral image, we get the activation of the four purple edges and the four green edges (for left, right, top, bottom borders). The \emph{gradients} for the four directions are then be approximated using subtraction. The box score is the smallest value among the four approximate gradients. To get the final proposals, we apply non-maximum suppression on all the boxes from all class activation maps (we show only the ``giraffe'' activation map here).}
    \label{fig:box_score}
\end{figure}

We consider each input box, $b(\bm{x}_i)_j$ as a candidate \emph{pseudo ground-truth} box for image $\bm{x}$; below we abbreviate the notation as $b_j, j \in \{1, \dots, B\}$, where $B$ is the number of candidate boxes. 
To rank boxes by their likelihood of being a true bounding box, we examine gradients along the edges of each candidate box, using the maps computed in Sec.~\ref{sec:multimodal-model}. 
We require that a bounding box should have \emph{strong gradients along all edges}.
Fig.~\ref{fig:box_score} shows the general idea of computing the box objectness.

Given $h(\bm{x}, c)$ which is the class activation map of a specific word/class $c$, we use the integral image technique \cite{viola2004robust} to compute $Act(b_i, c)$ which is the average activation score of class $c$ within proposal box $b_i$. 
We then use Eq.~\ref{eq:box_grad_left} to approximate the \emph{gradient} $g^l(b_i, c)$ along the left border of $b_i$. 
\begin{equation} \label{eq:box_grad_left}
    g^l(b_i, c) = Act(b_i^{l}, c) - Act(b_i^{l'}, c)
\end{equation}
In the formula, $b_i^{l}$ and $b_i^{l'}$ are the regions inside and outside the box accordingly (i.e., they denote the purple and green area in Fig.~\ref{fig:box_score}). We decide the size of these inside/outside regions by validating on the held-out validation set. We compute $g^r(b_i, c)$, $g^t(b_i, c)$, $g^b(b_i, c)$, which denote the right, top, and bottom gradients, in analogous fashion. 

The final objectness score of proposal $b_i$ is computed using Eq.~\ref{eq:box_score}, where $\beta$ is a hyper parameter to mediate the magnitude (how strong is the class activation in the region) and the gradients (how sharp are the edges), for any class. 
\begin{equation} \label{eq:box_score}
    O(b_i) = \max_c\left\{\beta \cdot Act(b_i, c) + \min_{k=l,r,t,b}(g^k(b_i, c))\right\}
\end{equation}
A natural explanation of Eq.~\ref{eq:box_score} is that the box is considered salient if all the four edges are sharp and most pixels inside it are active. 
A box is salient if it is detected in any of the class-aware activation maps, i.e. if at least one class produces a strong box. 

After getting the box objectness score $O(b_i) (i=1...B)$, we use non-maximum suppression, then set top-k of them as the pseudo ground-truth boxes.

\vspace{-0.3cm}
\paragraph{Differences with prior work.}

\begin{figure}[t]
    \centering
    \begin{subfigure}[b]{0.3\linewidth}
        \centering
        \includegraphics[width=1.0\linewidth]{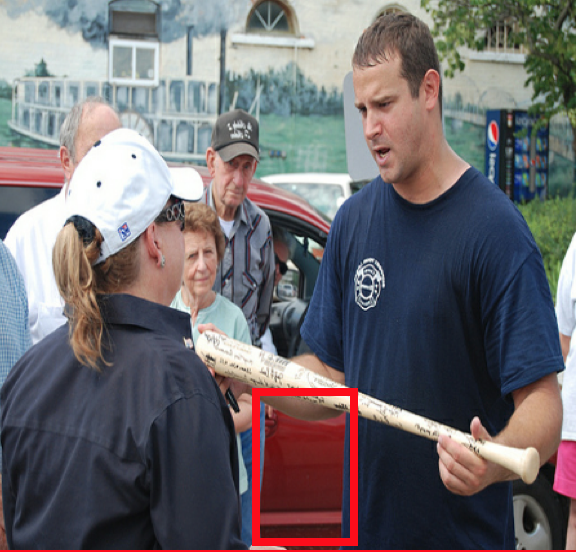}
        \caption{EdgeBoxes}
        \label{fig:weakness_edgeboxes}
    \end{subfigure}
    \begin{subfigure}[b]{0.3\linewidth}
        \centering
        \includegraphics[width=1.0\linewidth]{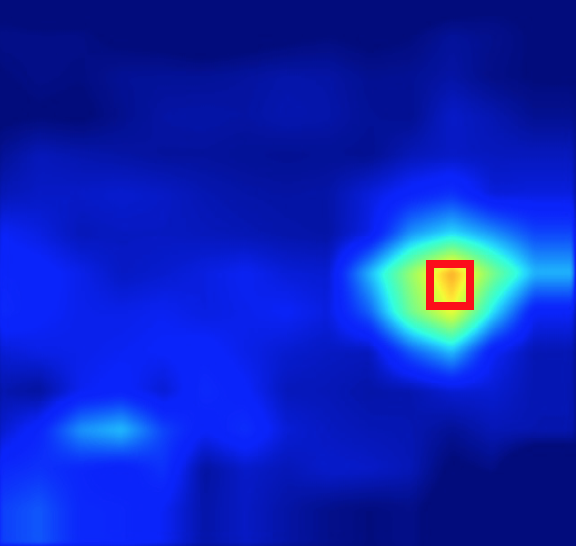}
        \caption{Diba et al.}
        \label{fig:weakness_diba}
    \end{subfigure}
    \begin{subfigure}[b]{0.3\linewidth}
        \centering
        \includegraphics[width=1.0\linewidth]{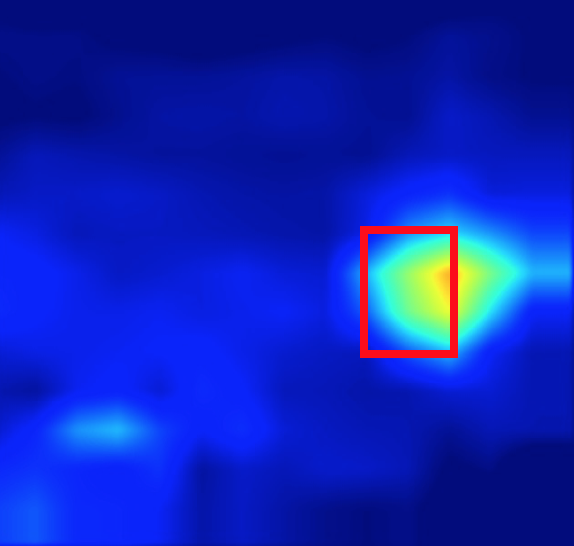}
        \caption{Wei et al.}
        \label{fig:weakness_wei}
    \end{subfigure}
    \caption{\textbf{The potential issues of previous methods}. The EdgeBoxes benefits the enclosed boxes formed by strong edges. Diba et al. \cite{Diba_2017_CVPR} is sub-optimal because it considers only the average activation inside the candidate box. Wel et al. \cite{Wei_2018_ECCV} proposed ``completeness'', which is the difference between the average activations inside and outside the box, yet it did not completely consider the situation that a box have sharp edges on three directions except the last one.}
    \label{fig:weakness}
\end{figure}

In contrast to prior work, we use a more strict criterion for objectness.
 \cite{Diba_2017_CVPR} only consider the magnitude (see Fig.~\ref{fig:weakness_diba}), while our method also considers the sharpness of the edges.
Compared to \cite{Wei_2018_ECCV} (see Fig~\ref{fig:weakness_wei}), we move one step further to also consider the effect of all of the four edges. 
Another difference is that the class labels $c$ in our case is not restricted to pre-defined object vocabulary (e.g. 80 in the case of COCO) since we can use any mined words from the caption vocabulary. 
The strict criterion helps obtain accurate boxes in this noisy setting.

\subsection{Associating boxes and class names}
\label{sec:mil}

\begin{figure}[t]
    \centering
    \includegraphics[width=1.0\linewidth]{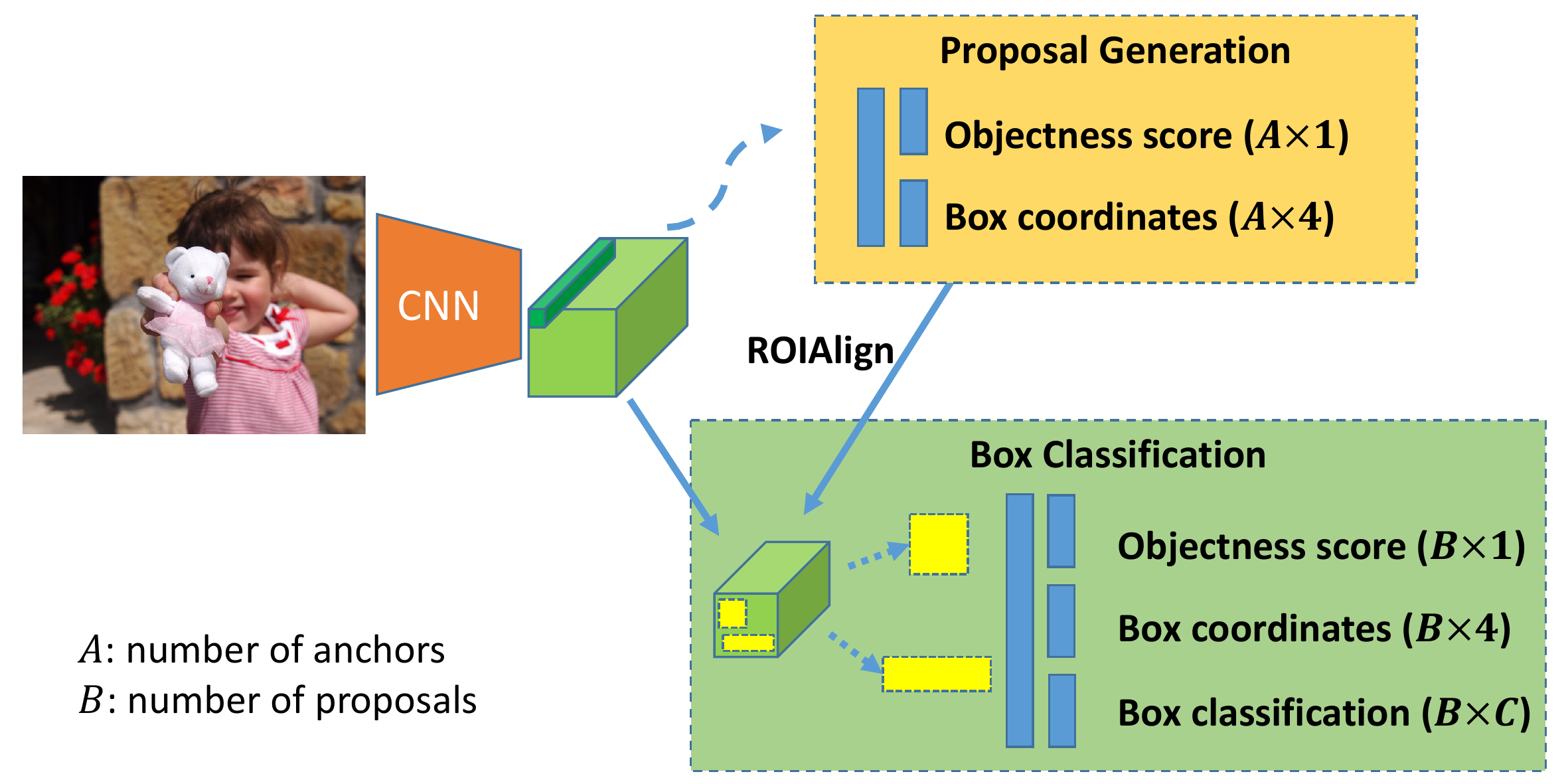}
    \caption{\textbf{Our modified Faster-RCNN model}.  Instead of directly predicting the $C + 1$ classes (including the background) in the box classifier, we factorize the 1-D objectness score and the $C$-way class indicator. This factorization is necessary in that the objectness score rejecting background is a box-level supervision and the multiple instance learning loss is a image-level supervision. Another difference from the Faster-RCNN model is that we use pseudo ground-truth boxes since we do not have the true location of boxes.
    }
    \label{fig:frcnn}
\end{figure}

Given the pseudo ground-truth boxes generated in Sec.~\ref{sec:pseudo_gt} and the image-level captioning annotations, we train a modified Faster-RCNN pipeline (see Fig.~\ref{fig:frcnn}). We use the pseudo ground-truth boxes to constraint the model to learn better objectness scores and box encodings, both in the ``proposal generation'' and the ``box classification'' stages. For assigning the semantic meaning, we use the multiple instance learning (MIL) similar to \cite{Diba_2017_CVPR, Wei_2018_ECCV}. 

The main difference between our modified model and the Faster-RCNN lies in the ``box classification'' stage. The final layer of our box classifier is a multi-head prediction layer, predicting a $1$-D box objectness score, a $4$-D box coordinates vector and a $C$-way class indicator. The pseudo ground-truth boxes (box-level supervision) are used to constrain the box objectness score and the box coordinates, while the multiple instance learning technique is used to guide the learning of box classification (image-level supervision). 
Non-maximum suppression is applied using the box objectness scores compared to Faster-RCNN that use the $1+C$ (including background) classification scores.

We next provide additional detail on MIL, assuming we have a bag of instances and their associated predictions. We first prune these boxes using the pseudo ground-truth and keep only the ones with IOU greater than $0.5$, resulting in matched boxes $p_i,i=1,\dots P$ and their associated predictions $f_{ic} \in \mathbb{R}^{P \times C}$ denoting the class membership of the boxes. We do not require them each to predict the target individually. Instead, we need only the most ``confident'' box to be responsible to the class label prediction (See Eq.~\ref{eq:mil_classification}). The procedure is somewhat similar to Global Maximum Pooling. The Eq.~\ref{eq:mil_loss} denotes the loss used for multiple instance learning where $y_c$ denotes the $C$-dimensional label vector extracted from image-level caption annotation; we normalize to ensure $\sum_c y_c =1$.
The most simple solution to extract the labels from image-level captions is to do the exact text match. Thus, $c$ would be considered a label if it appears in the caption. 

\vspace{-0.5cm}
\begin{equation}
\begin{split} \label{eq:mil_classification}
    P_c(\bm{x}, p_1 \cdots p_P)=\frac{\exp(\max_i f_{ic})}{\sum_{k=1}^C\exp(\max_j f_{jk}))}
\end{split}
\end{equation}
\vspace{-0.5cm}
\begin{equation}
\begin{split} \label{eq:mil_loss}
    L_{MIL}= -\sum_{c=1}^{C} &y_c \log (P_c(\bm{x}, p_1 \cdots p_P))
\end{split}
\end{equation}

\section{Experiments}
\label{sec:results}

We first describe our experimental setup and implementation details. 
We then show the quality of the reranked boxes we obtain, in relation to the boxes that are output by other objectness criteria. 
We also show how well we can retrieve words mined from captions, that match the object category vocabulary included with a given dataset.
Next, we show the accuracy of object detection in our weakly supervised learning setting. 
We also include a variety of qualitative results illustrating the different advantages of our model.

\subsection{Experimental setup}
\label{sec:metrics}

We conduct our main experiments on COCO \cite{lin2014microsoft, chen2015microsoft} which contains 82,783 training images, 40,504 validation images, and 40,775 test images. Caption annotations are available and contain 414,113 samples for training, 202,654 for validation.
Bounding box annotations are available for the same images, and we use them for validation. In all of our experiments, we use the images and captions from the training split for training, and report numbers on the held-out validation set or testing set.





\subsection{Implementation details}
\label{sec:details}

We implement all of the components in the paper using the TensorFlow \cite{Abadi:2016:TSL:3026877.3026899} framework.
To compute the multi-modal activation map, we use the MobilenetV1 \cite{howard2017mobilenets} as the backbone of the CNN network. We feed $448\times 448 \times 3$ images as input, and use the $14\times 14\times 1024$ output from "Conv2d\_13\_pointwise" layer. 
The input image is separated into $14\times 14$ regions, and each is represented by a $1024$-D vector. The parameters in the MobilenetV1 are frozen during training. For the word embeddings, we use $50D$ vectors to represent the words in the caption vocabulary $V$, initialized from the GloVe \cite{pennington2014glove} model and updated during the training process. For the convolutional layers and fully connected layers, we use batch normalization \cite{ioffe2015batch} with decay of $0.999$. The margin of the triplet loss is set to $0.1$ in Eq.~\ref{eq:triplet_loss}. For the triplet loss optimization (Sec.~\ref{sec:multimodal-model}), we use the Adam optimizer \cite{kingma2014adam} with a learning rate of $0.003$.
The model is trained for $200,000$ steps with a batch size of $32$ (roughly 75 epochs), and the top-1 accuracy of the text retrieval (in-batch retrieval for around 200 captions per image) is around $85\%$. We observe that using $224\times 224$ image provides better results, but the resulting $7\times 7 \times 1024$ output is much coarser.

\begin{figure*}[t]
    \centering
    \includegraphics[width=0.33\linewidth]{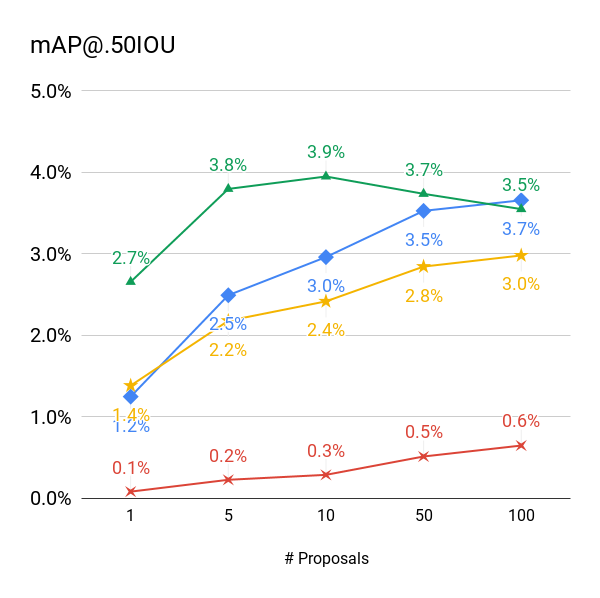}
    \includegraphics[width=0.33\linewidth]{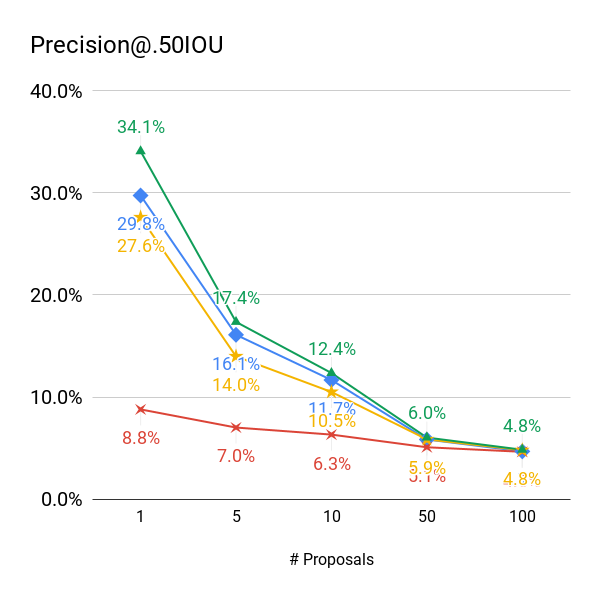}
    \includegraphics[width=0.33\linewidth]{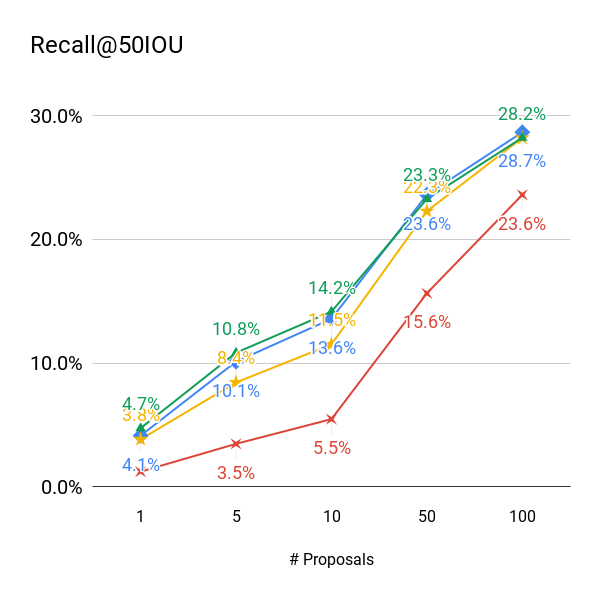}
    \centering
    \includegraphics[width=0.4\linewidth]{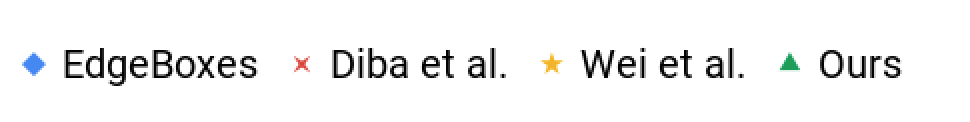}
    \caption{\textbf{The quality of different pseudo ground-truth boxes (COCO)}. We compare the performance of the boxes proposed by EdgeBoxes \cite{zitnick2014edge}, Diba et al. \cite{Diba_2017_CVPR}, Wei et al. \cite{Wei_2018_ECCV}, and the ones generated using our Eq.~\ref{eq:box_score}.}
    \label{fig:quality_gt}
\end{figure*}

To generate the final set of pseudo ground-truth boxes, we run the EdgeBoxes~\cite{zitnick2014edge} to provide the proposals and use Eq.~\ref{eq:box_score} to rank the candidates based on the learned class activation map. We choose $2\%$ of the box size as a margin to compute the approximated gradients in Eq.~\ref{eq:box_grad_left} and set $\beta$ to be $0.005$ in Eq.~\ref{eq:box_score} as we validated these two parameters on the COCO validation set.

\subsection{Quality of proposed boxes}

We evaluate the quality of our pseudo ground-truth boxes. (Sec.~\ref{sec:pseudo_gt}, Eq.~\ref{eq:box_score}) shows a comparison of our result to others:
\begin{itemize}[nolistsep,noitemsep]
    \item EdgeBoxes \cite{zitnick2014edge}: a contour image based baseline implemented using OpenCV \cite{opencv_library} with default parameters.
    \item our ablation of Diba et al. \cite{Diba_2017_CVPR} which uses our framework but considers only the average activation inside the box.
    \item our abltion of Wei et al. \cite{Wei_2018_ECCV} which considers the difference of the activation inside and outside the box. We experiment to use border width of $0.05$, $0.1$, and $0.2$ and report the best performance of using border width $0.2$.
\end{itemize}

All methods rank the top-$300$ boxes proposed by EdgeBoxes.
For reference, the mAP@0.5IOU, Precision@0.5IOU and Recall@0.5IOU of the top-1 proposal generated by the Faster-RCNN model trained on COCO are $10.1\%$, $78.6\%$, and $10.9\%$ respectively.

We use the $80$ objects in the COCO vocabulary in Eq.~\ref{eq:box_score} in order to enable quantitative evaluation, though our method has the ability to discover the vocabulary. 
Metrics reported are mean Average Precision (mAP), Precision, Recall and we evaluate top-1, top-5, top-10, top-50, and top-100 boxes. The proposed boxes are treated as correct if they have an IOU greater than $0.5$ with any annotated boxes.



In Fig.~\ref{fig:quality_gt}, we observe that when up to 50 proposals are considered, our method significantly outperforms all baseline methods. When 100 proposals are generated, our method's performance is comparable to that of EdgeBoxes, and stronger than the other methods. Our method's key strength lies in generating very precise boxes. 
To better understand the relationship of our method to the baselines, note that the EdgeBoxes model considers only the aggregation of edges while ignoring the semantic meaning of each image pixel. Wei et al. make an argument that Diba et al.'s approach is sub-optimal in that it has bias toward smaller boxes with higher class activation. Wei et al. have greatly improved Diba et al.'s approach, but the ``completeness'' criterion they propose (which encourages their model to generously include a large set of high-confidence pixels into the box) results in including inaccurate boxes that have only three strong edges. In Fig.~\ref{fig:weakness} we show cases where this criterion results in problematic results.
Note that none of the baseline methods were previously compared in the challenging setting of COCO. 


\subsection{Quality of the learned detection model}

 Tab.~\ref{tab:main} gives a quantitative result of discovering and localizing visual objects. We see our method achieves $~7\%$ mAP@IoU=0.50. Considering the extremely low probability of guessing a box correctly by chance, and the very low probability of assigning a correct label to a box by  chance, the $~7\%$ mAP we got with a model trained solely from noisy captions proves that our method does manage to learn useful localized visual concepts. Moreover, we use ``harsh'' string matching to count whether a detected object class is correct or not. ``Gentleman", ``lady", or ``baby" would be considered as a completely wrong prediction when the ground truth label is ``person". While captions written by different annotators are quite diverse (see Fig.~\ref{fig:concept}), object labels (which other methods require) are standardized, thus WSOD methods do not have to handle synonymy. 

We are not aware of any other methods that learn detection on COCO in a weakly supervised manner, especially from captions. However, weakly supervised methods on PASCAL VOC are roughly only 50\% as accurate as supervised methods \cite{Wei_2018_ECCV,redmon2016you}. COCO is a more complex dataset than PASCAL, thus we expect a much larger gap between supervised and weakly supervised methods such as ours.

 

\begin{table}[t]
    \centering
    \begin{tabularx}{\linewidth}{|>{\hsize=1.3\hsize}s|>{\hsize=0.7\hsize}s|}
        \hline
         & Our Method \\
        \hline \hline
        mAP & 0.0241 \\
        mAP@0.50IOU & 0.0691 \\
        \hline
        AR@1 & 0.0369 \\
        AR@10 & 0.0656 \\
        AR@100 & 0.0683 \\
        \hline
    \end{tabularx}
    \caption{\textbf{Detection results on MS COCO.}}
    \label{tab:main}
\end{table}

\subsection{Quality of the learned vocabulary}

\begin{figure}[t]
    \centering
    \includegraphics[width=0.8\linewidth]{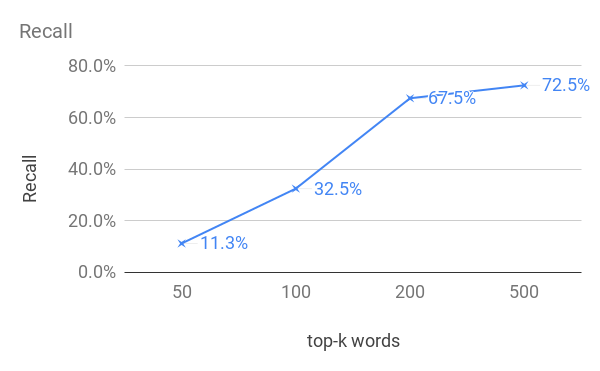}
    \caption{\textbf{The quality of the learned vocabulary}. We rank the captioning vocabulary using Eq.\ref{eq:mining_objects} and filter out less mentioned words. The recall scores of the $80$ COCO categories are reported.}
    \label{fig:coco_label_recall}
\end{figure}

We want to see if the top-ranked words in vocabulary $C$ are actually the words related to the objects. Therefore, we design the following experiment that rank the words in the captioning vocabulary $V$ using Eq.\ref{eq:mining_objects} and use the top-k ($k=50,100,200,500$) recall as a measurement to see how well the top-k candidates can retrieve the $80$ COCO labels. In order to match the captioning words to the compound words in COCO labels, we manually designate that ``baseball glove'' can be matched by ``glove'', ``baseball bat'' can be matched by ``bat'', ``tennis racket'' can be matched by ``racket'', and so on. Beyond using Eq.\ref{eq:mining_objects}, we also use the word frequency to prune words mentioned less than $200$ times. We do not process further optimization such as utilizing the morphology features, merging synonyms and so on. 
A summary result that measures recall of COCO words using our mining strategy is shown in Fig.~\ref{fig:coco_label_recall}.

\section{Conclusion}
We proposed a method which managed to learn the relationship between image pixels with words from free form text. 
We achieved this by learning from noisy captions only.
This opens the possibility of training an object detector on massive and readily available internet image-text data.

There are still many directions to explore. Firstly, we will examine to what extent the multiple resolution information used in weakly supervised work \cite{ronneberger2015u, Zhu_2015_CVPR,Diba_2017_CVPR, Wei_2018_ECCV} helps in our setting. Secondly, the multi-modal information of the edges \cite{zitnick2014edge} and the class activation maps can be fused in that they both rely on integral image based computations. We will also explore running edge boxes on class activation maps and then aggregating the results. Thirdly, we will extend our approach and train it end-to-end, sharing the basic convolutional layers. 
We believe these extensions will even more strongly highlight the promise of learning in an open-vocabulary setting from the diverse resources of the web.


\paragraph{Supplementary Material}
More experiments, result visualizations and extensive analysis can be found in the appendices (attached after references), including:
\begin{enumerate}
    \itemsep0em 
    \item Visualization of the class-aware activation map
    \item Visualization of the generated pseudo ground-truth boxes
    \item Quantitative results on Pascal VOC
    \item Sample results on novel objects discovery
    \item Evaluation of COCO-trained models on PASCAL VOC dataset
\end{enumerate}

{\small
\bibliographystyle{ieee}
\bibliography{egbib}
}

\onecolumn
\appendix 
\renewcommand{\thesection}{Appendix \Alph{section}}

\setcounter{section}{0}

\newpage

\section{Visualization of the class-aware activation map}
\label{sec:supp_vis_cam}

\begin{figure}[h]
    \centering
    \includegraphics[width=0.9\linewidth]{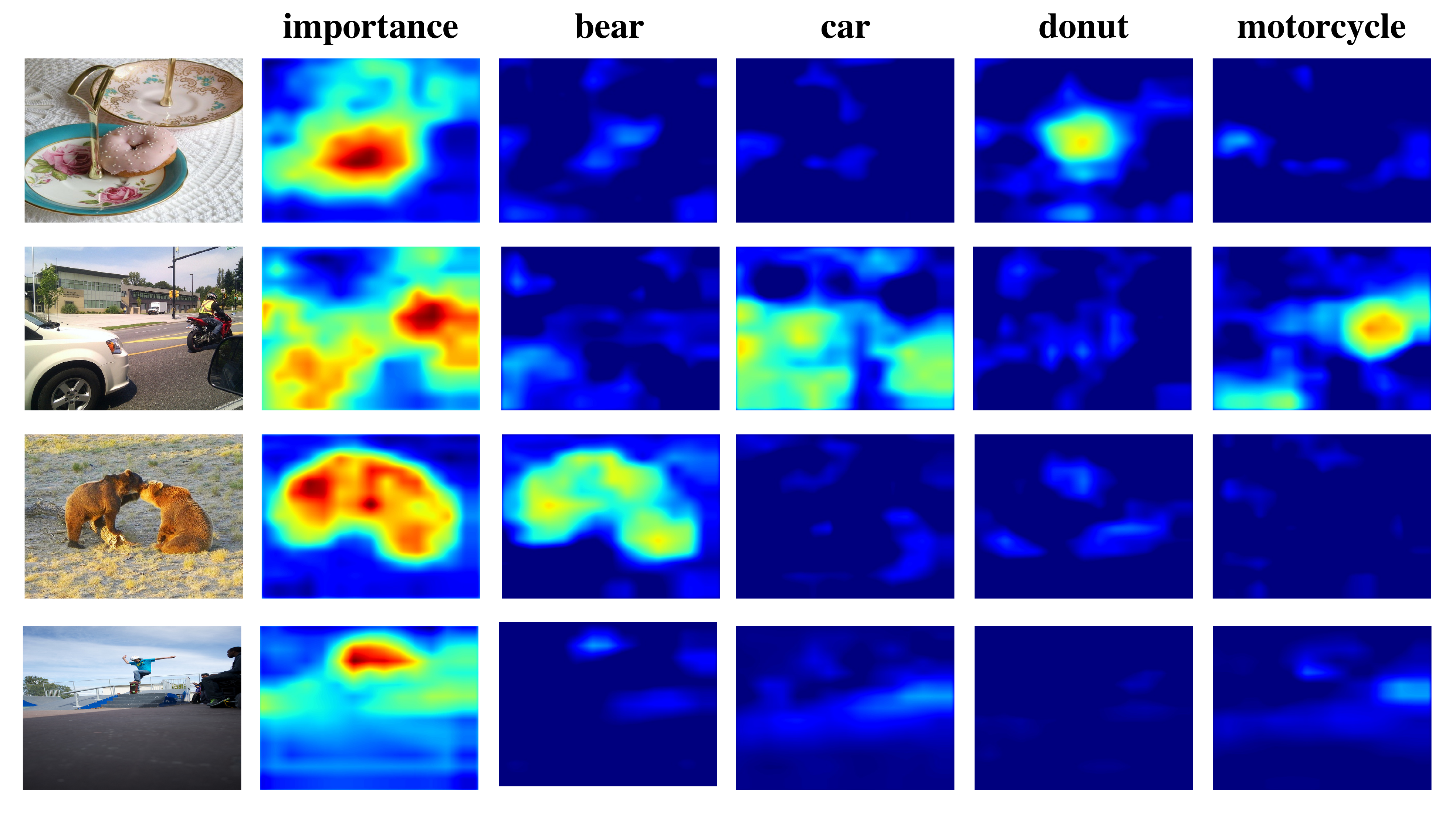}
    \caption{\textbf{The learned class activation maps}. For visualization purpose, the values for image importance score $S^{img}$ (the second column) are normalized 
    to be within [0, 1].
    However, for the class-aware activation maps $h(\bm{x}, c)$ 
    (the third till the last column), we directly show the heat map for the original values (no normalization).}
    \label{fig:vis_act_map}
\end{figure}

We demonstrate the learned class activation map in Fig.~\ref{fig:vis_act_map}, in which the first column shows the original image, the second column shows the image importance score obtained from Eq.~3 in the main text, and the rest columns each shows a class-aware activation map from Eq.~6. We can clearly see from the heatmap that the class activation map focuses only on the image regions that are semantically related to the query words (we only show ``bear'', ``car'', ``donut'', and ``motorcycle'' for demonstration purpose). 
It is worth noting that the activation map can be generated using any word from the vocabulary of caption.

 \newpage

\section{Visualization of the generated pseudo ground-truth boxes}
\label{sec:supp_vis_pseudo_gt}

\begin{figure}[h]
    \centering
    \includegraphics[width=0.9\linewidth]{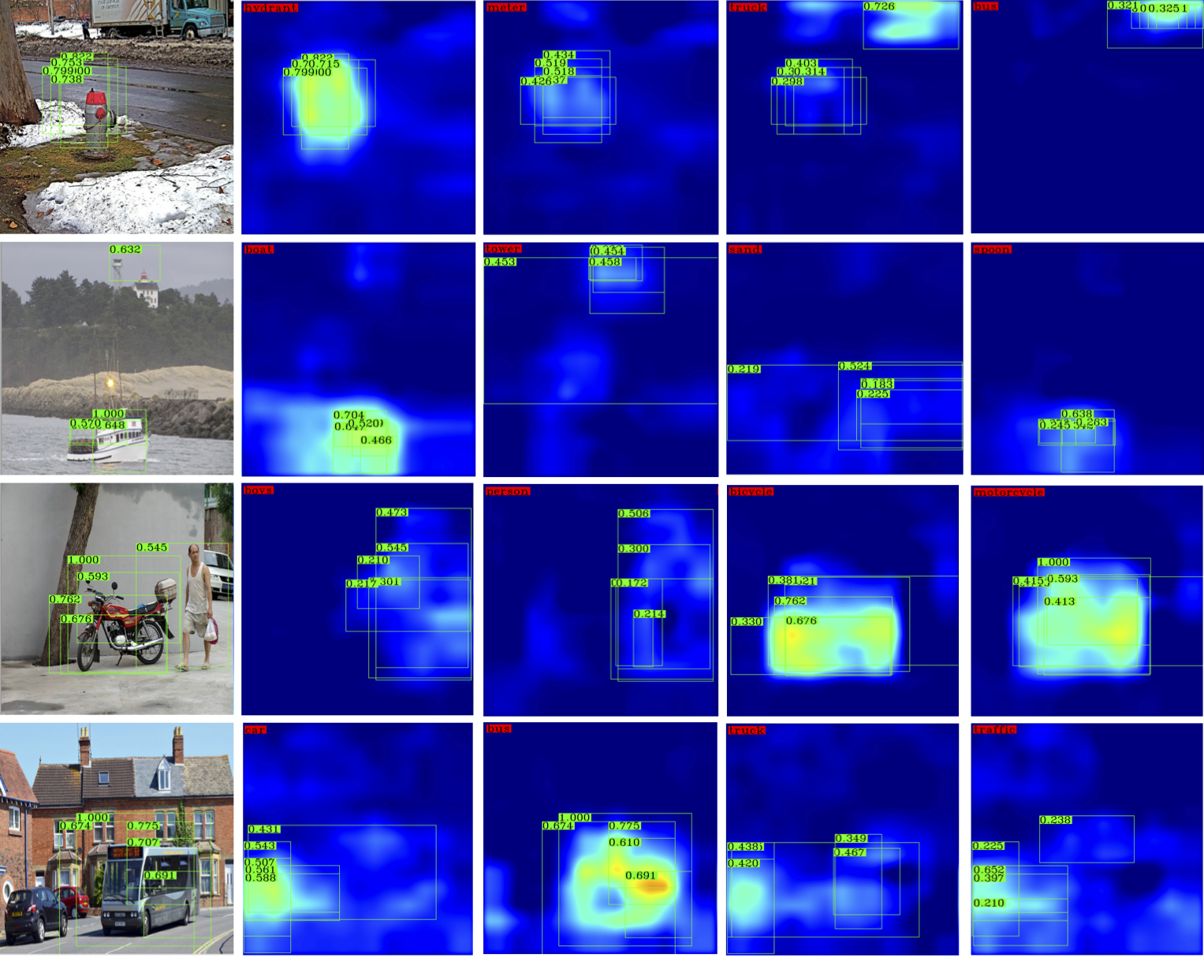}
    \caption{\textbf{The box selection procedure}. We show images with top-3 pseudo ground-truth overlaid on the image, in the first column. The remaining columns show the potential pseudo ground-truth boxes of each class.}
    \label{fig:vis_pseudo_gt}
\end{figure}

 Given the candidate boxes, we compute the box objectness score (Eq.~9) based on the integral images of all class channels. Then the normal non-maximum suppression (NMS) procedure is applied to all the boxes and all the class activation maps. We show this NMS procedure in Fig.~\ref{fig:vis_pseudo_gt} where the second till the last column shows the boxes with the highest scores in typical class activation map and the first column shows the final aggregated boxes.

 \newpage

\section{Quantitative results on Pascal VOC 2007}
\label{sec:supp_quati_voc}

We provide experiments on the Pascal VOC 2007 dataset [12] to provide additional comparison to methods from the weakly supervised object detection literature. In order to make the training data analogous to the COCO data, we concatenate the labels within an example to form a caption, and use the same triplet loss based technique as that in Sec. 3.1 of the main text. On this dataset, we use the \emph{trainval} set (5,011 images) of VOC 2007 and the \emph{trainval} set (11,540) of VOC 2012 for training, and report number on the VOC 2007 \emph{test} set (4,952 images).

\paragraph{Quality of the pseudo ground-truth}

\begin{figure}[h]
    \centering
    \includegraphics[width=0.33\linewidth]{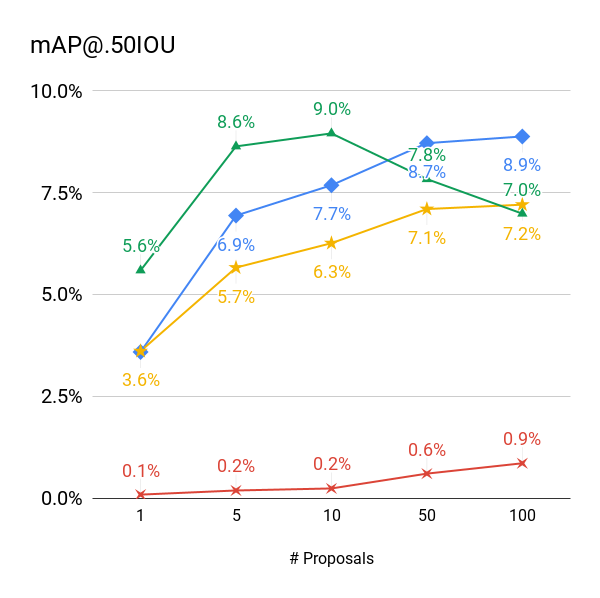}
    \includegraphics[width=0.33\linewidth]{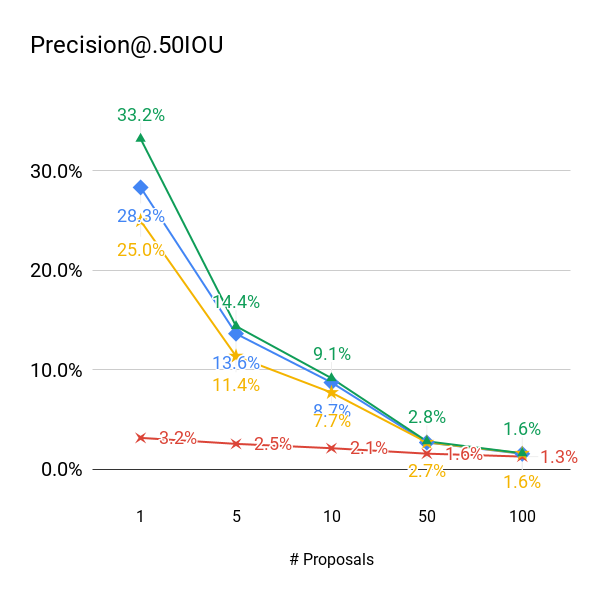}
    \includegraphics[width=0.33\linewidth]{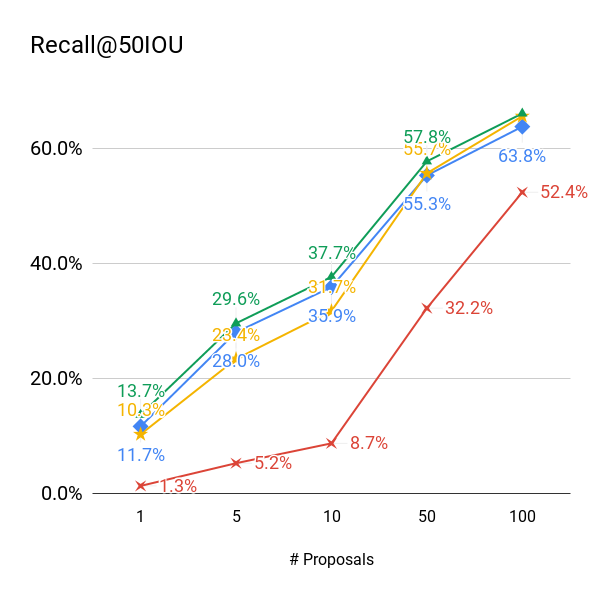}
    \centering
    \includegraphics[width=0.4\linewidth]{figs/rpn_class_map_legend.png}
    \caption{\textbf{The quality of different pseudo ground-truth boxes (VOC 2007)}. We compare the performance of the boxes proposed by EdgeBoxes~[45], Diba et al.~[10], Wei et al.~[41], and the ones generated using our Eq.9.}
    \label{fig:quality_gt_voc}
\end{figure}

We first compare the generated pseudo ground-truth to the baseline methods of EdgeBoxes~[45], Diba et al.~[10], and Wei et al.~[41]. The basic setting is the same as the in. 4.3 in our paper where we evaluate class-agnostic box locations but the evaluation is processed on the Pascal VOC instead of the COCO data. In Fig.~\ref{fig:quality_gt_voc} we show the evaluation metrics of mAP@0.5IOU, Precision@0.5IOU, and Recall@0.5IOU. Similar to what we observed on COCO, we find that when up to 10 proposals are considered, our method significantly outperforms all baseline methods and the key strength still lies in precision. To conclude, our method provides the most precise boxes among the four methods on Pascal VOC.

\paragraph{Detection results}

To further investigate the impact of the pseudo ground-truth boxes, we compare our method to the baseline methods by evaluating the trained modified Faster-RCNN pipeline (Sec. 3.3). We use the same class-activation map learned by our multi-modal model (Sec. 3.1), thus these methods only differ in the box objectness function. The results are shown in Tab.~\ref{tab:pascal_voc}. Although it's not an end-to-end comparison, \textbf{we observe that our proposed objectness function outperforms other objectness criteria from both of these prior methods.} 

\begin{table}[h]
    \footnotesize
    \centering
    \caption{\textbf{Comparison of detection average precision (AP) (\%) on PASCAL VOC.} Please note that Diba et al. and Wei et al. report better results in their own papers since we here compare only the box objectness function while keeping the class activation map the same. Our learned class activation map is inferior in that it is not fine-tuned using an additional segmentation stage. The best per-category score is highlighted in bold.}
    \begin{tabularx}{\textwidth}{|>{\hsize=3\hsize}s|*{19}{>{\hsize=0.85\hsize}s}>{\hsize=1.1\hsize}s|>{\hsize=1.75\hsize}s|}
    \hline
        methods & \rotatebox{90}{aero} &\rotatebox{90}{bike} & \rotatebox{90}{bird} & \rotatebox{90}{boat} & \rotatebox{90}{bottle} & \rotatebox{90}{bus} & \rotatebox{90}{car} & \rotatebox{90}{cat} & \rotatebox{90}{chair} & \rotatebox{90}{cow} & \rotatebox{90}{table} & \rotatebox{90}{dog} & \rotatebox{90}{horse} & \rotatebox{90}{mbike} & \rotatebox{90}{person} & \rotatebox{90}{plant} & \rotatebox{90}{sheep} & \rotatebox{90}{sofa} & \rotatebox{90}{train} & \rotatebox{90}{tv} & mAP \\
    \hline \hline
        Diba &  $8.4$ & $10.3$ & $15.8$ &  $7.2$ & $\mathbf{1.9}$ & $17.2$ & $12.7$ &  $6.8$ & $\mathbf{3.1}$ & $15.4$ &  $3.2$ &  $6.3$ & $10.3$ & $15.5$ & $5.6$ & $4.6$ & $\mathbf{22.9}$ &  $3.3$ & $16.3$ & $\mathbf{35.6}$ & $11.1$ \\
        Wei  & $\mathbf{38.5}$ & $\mathbf{29.5}$ & $\mathbf{26.9}$ & $\mathbf{17.1}$ & $1.6$ & $\mathbf{46.1}$ & $31.5$ & $42.6$ & $1.4$ & $20.3$ & $10.4$ & $40.7$ & $24.1$ & $32.3$ & $9.8$ & $5.2$ & $18.3$ & $\mathbf{17.5}$ & $51.1$ & $24.0$ & $24.4$ \\
    \hline
        Ours & $37.0$ & $28.4$ & $26.6$ & $14.0$ & $1.3$ & $43.7$ & $\mathbf{34.4}$ & $\mathbf{48.3}$ & $1.2$ & $\mathbf{20.5}$ & $\mathbf{11.8}$ & $\mathbf{40.9}$ & $\mathbf{31.8}$ & $\mathbf{33.9}$ & $\mathbf{9.9}$ & $\mathbf{7.0}$ & $19.4$ & $15.3$ & $\mathbf{53.5}$ & $22.5$ & $\mathbf{25.1}$ \\
    \hline
    \end{tabularx}
    \label{tab:pascal_voc}
\end{table}


 \newpage

\section{Discovering novel objects}
\label{sec:novel_objects}

\paragraph{Vocabulary of novel objects} Though we design a weakly supervised object detection model for discovering novel objects, all of our experiments in the paper are built based on the pre-defined $80$ COCO objects vocabulary. In this experiment, we use the Eq.~7 to discover objects beyond the $80$ classes in the COCO dataset. We first sort the words in the captioning vocabulary $V$ (4526 words) by their word importance scores (obtained form Eq.~4) and keep only the top-1000 of them. Then we filter out infrequent words (mentioned less than 1000 times) and the words too similar to the $80$ COCO labels (if their cosine similarities are greater than $0.6$). The resulting discovered vocabulary is shown in Table \ref{tab:novel_words}.

\begin{table}[h]
    \centering
    \caption{\textbf{Discovered novel words beyond the $\mathbf{80}$ COCO labels}. We use the mentioned method (Eq.~7) to discover interesting objects beyond the $80$ COCO categories. The columns we show are the discovered words, word frequency in the captioning training set, the word importance score, the most similar COCO label word, and the cosine similarity to the most similar COCO label word. The higher word importance score indicates more distinguishable image feature associated with the word. }
    \label{tab:novel_words}
    \begin{tabular}{|c|c|c|c|c|}
    \hline
    word & freq & score & coco word & similarity \\
    \hline
    \hline
    soccer&1965&3.71&frisbee&0.43\\
    shower&1602&3.11&sink&0.48\\
    salad&1034&2.44&sandwich&0.45\\
    beach&7953&2.41&kite&0.32\\
    herd&2268&2.17&sheep&0.48\\
    kitchen&9366&2.12&oven&0.59\\
    signs&2684&2.12&stop&0.47\\
    bridge&1275&2.10&boat&0.36\\
    children&2248&1.99&skis&0.21\\
    baby&3174&1.98&carrot&0.35\\
    \hline
    \end{tabular}
    \quad
    \begin{tabular}{|c|c|c|c|c|}
    \hline
    word & freq & score & coco word & similarity \\
    \hline
    \hline
    mirror&3572&1.93&sink&0.56\\
    animals&1987&1.92&sheep&0.49\\
    boys&1170&1.90&person&0.25\\
    sand&1264&1.90&kite&0.31\\
    tower&2957&1.89&clock&0.39\\
    girls&1139&1.88&person&0.25\\
    cabinets&1196&1.86&dryer&0.50\\
    child&4329&1.84&person&0.32\\
    zoo&1319&1.84&giraffe&0.48\\
    kids&1253&1.81&train&0.20\\
    \hline
    \end{tabular}
\end{table}


\paragraph{Visualization of the detected novel objects}

\begin{figure}[h]
    \centering
    \includegraphics[width=0.8\linewidth]{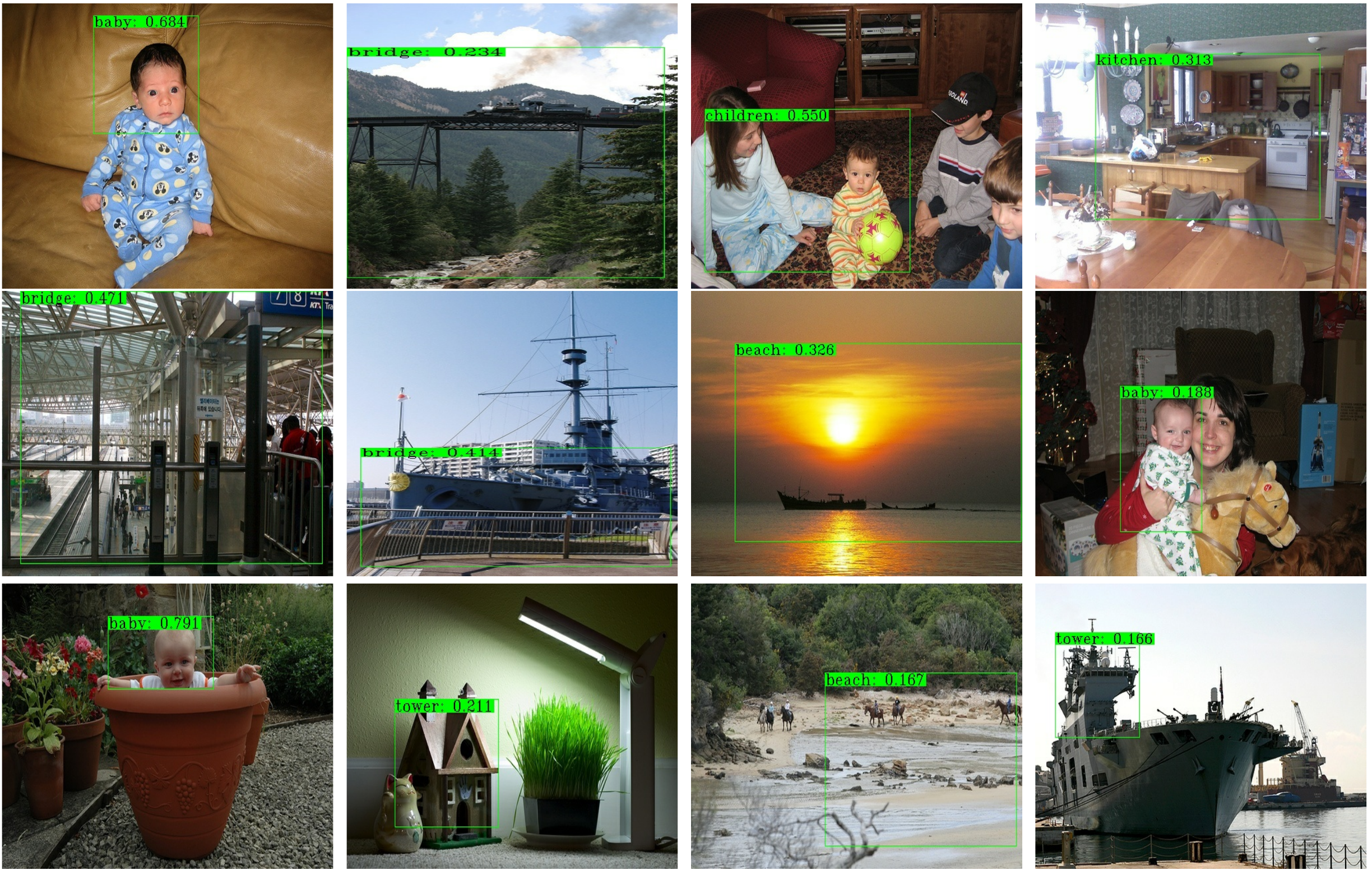}
    \caption{\textbf{Qualitative results of the detected novel objects}.}
    \label{fig:vis_novel_objects}
\end{figure}

We show qualitative results for the detected novel objects in Fig.~\ref{fig:vis_novel_objects}.

 \newpage

\section{Evaluating the COCO-trained model on Pascal VOC}
\label{sec:supp_coco_on_voc}

In this section, we investigate the performance of our model trained on captions from COCO and evaluated on Pascal VOC 2007.
This setting is most close to the real world problems that we proposed to solve: learning to discover and detect objects from the abundant and cost-free caption data. 
It is an extremely difficult problem since captions are only loosely related to the visual objects in images.
As shown in Tab.~\ref{tab:coco_voc}, our method achieved mAP of 24.8\% on Pascal.
It is not so much of a surprise that our performance are far behind from the number (45\%) reported by other weakly supervised method, which requires accurate, human annotated image level labels for the object detection task.
However, learning from the labor-free, and widely available image captions gives us an unparalleled advantage over others. 
The cost of using our model to discover and detect a novel visual object class is very little.
As a result, it is possible for our model to scale up to detect millions of object concepts.
Nevertheless, our model might be considered as an effective and efficient tool for data bootstrapping as well.
For 2 classes out of 20, ie. \textit{train} and \textit{buses}, we got very convincing result ($mAP > 50\%$).
This tells that it might not be necessary to label data for all classes from scratch. 

To sum up, learn to detect objects using only captions could open many possibilities for future application.
Our result confirms that our model is effective to utilize this data.
As far as we know, we are the first one who proposed a method that works in such highly challenging setting.

\begin{table}[h]
    \footnotesize
    \centering
    \caption{\textbf{Evaluate the COCO-trained model on the Pascal VOC data.} We show our model trained using COCO captioning data (82,783 images with 414,113 captions). The VOC model (trained on 16,551 images) is the same as the one described in Sec. 3 of this document. The evaluation is processed on the 4,952 test images of the Pascal VOC 2007 data. The best per-category score is highlighted in bold.}
    \begin{tabularx}{\textwidth}{|>{\hsize=2.5\hsize}s|*{20}{>{\hsize=0.85\hsize}s}|>{\hsize=2.5\hsize}s|}
    \hline
        train & \rotatebox{90}{aero} &\rotatebox{90}{bike} & \rotatebox{90}{bird} & \rotatebox{90}{boat} & \rotatebox{90}{bottle} & \rotatebox{90}{bus} & \rotatebox{90}{car} & \rotatebox{90}{cat} & \rotatebox{90}{chair} & \rotatebox{90}{cow} & \rotatebox{90}{table} & \rotatebox{90}{dog} & \rotatebox{90}{horse} & \rotatebox{90}{mbike} & \rotatebox{90}{person} & \rotatebox{90}{plant} & \rotatebox{90}{sheep} & \rotatebox{90}{sofa} & \rotatebox{90}{train} & \rotatebox{90}{tv} & mAP \\
    \hline \hline
        VOC & \textbf{37.0} & 28.4 & \textbf{26.6} & 14.0 & 1.3 & 43.7 & \textbf{34.4} & \textbf{48.3} & \textbf{1.2} & \textbf{20.5} & \textbf{11.8} & 40.9 & 31.8 & 33.9 & \textbf{9.9} & \textbf{7.0} & \textbf{19.4} & 15.3 & 53.5 & \textbf{22.5} & \textbf{25.1} \\
        COCO & 35.1 & \textbf{30.2} & 21.6 & \textbf{17.4} & \textbf{1.6} & \textbf{49.1} & 33.4 & 44.4 & 0.4 & 16.2 & 8.4 & \textbf{42.5} & \textbf{37.3} & \textbf{37.2} & 6.1 & 4.0 & 14.1 & \textbf{25.5} & \textbf{57.4} & 14.1 & 24.8 \\
    \hline
    \end{tabularx}
    \label{tab:coco_voc}
\end{table} \newpage

\end{document}